\documentclass[letterpaper, 10 pt, conference]{ieeeconf}  

\IEEEoverridecommandlockouts                              

\usepackage{times}
\usepackage{latexsym}
\usepackage{graphicx}
\usepackage{amsmath}
\usepackage{amssymb}
\usepackage{amsfonts}
\usepackage{booktabs}
\usepackage{bbm}
\usepackage{multicol}
\usepackage{algorithm}
\usepackage{algpseudocode}
\usepackage{hyperref}
\usepackage[dvipsnames]{xcolor}
\usepackage{float}
\usepackage{wrapfig}
\usepackage[export]{adjustbox}
\definecolor{darkgreen}{RGB}{0,128,0}
\usepackage{caption}
\usepackage{subcaption}
\usepackage{longtable} 
\usepackage{booktabs} 
\usepackage{tabularx} 
\usepackage{xspace}
\usepackage[capitalise]{cleveref}
\let\labelindent\relax
\usepackage[inline]{enumitem}
\usepackage{soul}
\usepackage{listings}
\usepackage{ltablex}
\usepackage{csquotes}
\usepackage{tcolorbox}
\usepackage{fancyvrb}
\usepackage{tcolorbox}
\tcbuselibrary{skins} 
\usepackage{caption}
\usepackage{newfloat}

\DeclareFloatingEnvironment[fileext=loq,placement={!ht},name=Prompt]{quotebox}
\DeclareFloatingEnvironment[fileext=loq,placement={!ht},name=LLM Output]{llmoutput}
\newtcolorbox{myquote}{colback=orange!5!white, colframe=orange!50!white, fonttitle=\bfseries, colbacktitle=red!85!black, enhanced,fontupper=\itshape}
\newtcolorbox{llm_quote}{colback=blue!5!white, colframe=blue!50!white, fonttitle=\bfseries, colbacktitle=red!85!black, enhanced,fontupper=\itshape}

\usepackage[backend=biber,
            hyperref=true,
            url=false,
            isbn=false,
            doi=false,
            backref=false,
            style=ieee,
            citestyle=numeric-comp,
            sorting=nyt,
            block=none]{biblatex}
\usepackage[font=footnotesize]{caption}
\newcommand{\methodname}{OLAF\xspace}

\newcommand{\hcil}{HIL}
\newcommand{\Hcil}{human-in-the-loop imitation learning}

\definecolor{Burgundy1}{RGB}{128,0,32}
\definecolor{cerulean}{rgb}{0.0, 0.48, 0.65}

\title{\LARGE \bf
Interactive Robot Learning from Verbal Correction
}

\author{Huihan Liu$^{1,2}$, Alice Chen$^{1}$, Yuke Zhu$^{2}$, Adith Swaminathan$^{1}$, Andrey Kolobov$^{1}$, Ching-An Cheng$^{1}$
\thanks{*This work was partially done during Huihan Liu's internship at Microsoft Research. Correspondence: \url{huihanl@utexas.edu}}
\thanks{$^{1}$Microsoft Research, Redmond, WA 98052, USA    
        {\tt\small \{adswamin,akolobov,chinganc\}@microsoft.com}}%
\thanks{$^{2}$The University of Texas at Austin, Austin, TX 78712, USA
        {\tt\small \{huihanl,yukez\}@utexas.edu }}%
}

\begin{document}

\maketitle

\begin{abstract}
The ability to learn and refine behavior after deployment has become ever more important for robots  as we design them to operate in unstructured environments like households.
In this work, we design a new learning system based on large language model (LLM), \methodname{}, that allows everyday users to teach a robot using verbal corrections when the robot makes mistakes, e.g., by saying ``Stop what you're doing. You should move closer to the cup."
A key feature of \methodname{} is its ability to \emph{update} the robot's visuomotor neural policy based on the verbal feedback to avoid repeating mistakes in the future. This is in contrast to existing LLM-based robotic systems, which only follow verbal commands or corrections but not learn from them.
We demonstrate the efficacy of our design in experiments where a user teaches a robot to perform long-horizon manipulation tasks both in simulation and on physical hardware, achieving on average 20.0\% improvement in policy success rate. Videos and more results are at \textcolor{cerulean}{\url{https://ut-austin-rpl.github.io/olaf/}}.

\end{abstract}

\section{Introduction}

Imagine training a robot by \emph{talking} to it about the ways it could have completed a given task better. This is similar to how a child learns when s/he spills water and is advised afterwards: ``To avoid this in the future, please try to keep your cup upright''. In this work, we show that, with the advent of large language models (LLMs) such as GPT-4~\cite{openai2023gpt4}, teaching a robot using  verbal corrections is now possible.
To that end, we introduce \textbf{\methodname{}} (\textbf{O}peration-relabeled Learning with \textbf{LA}nguage \textbf{F}eedback), a system that \emph{learns} robot policies interactively using verbal corrections.%

\begin{figure}[t]
    \centering
    \includegraphics[width=1\linewidth]{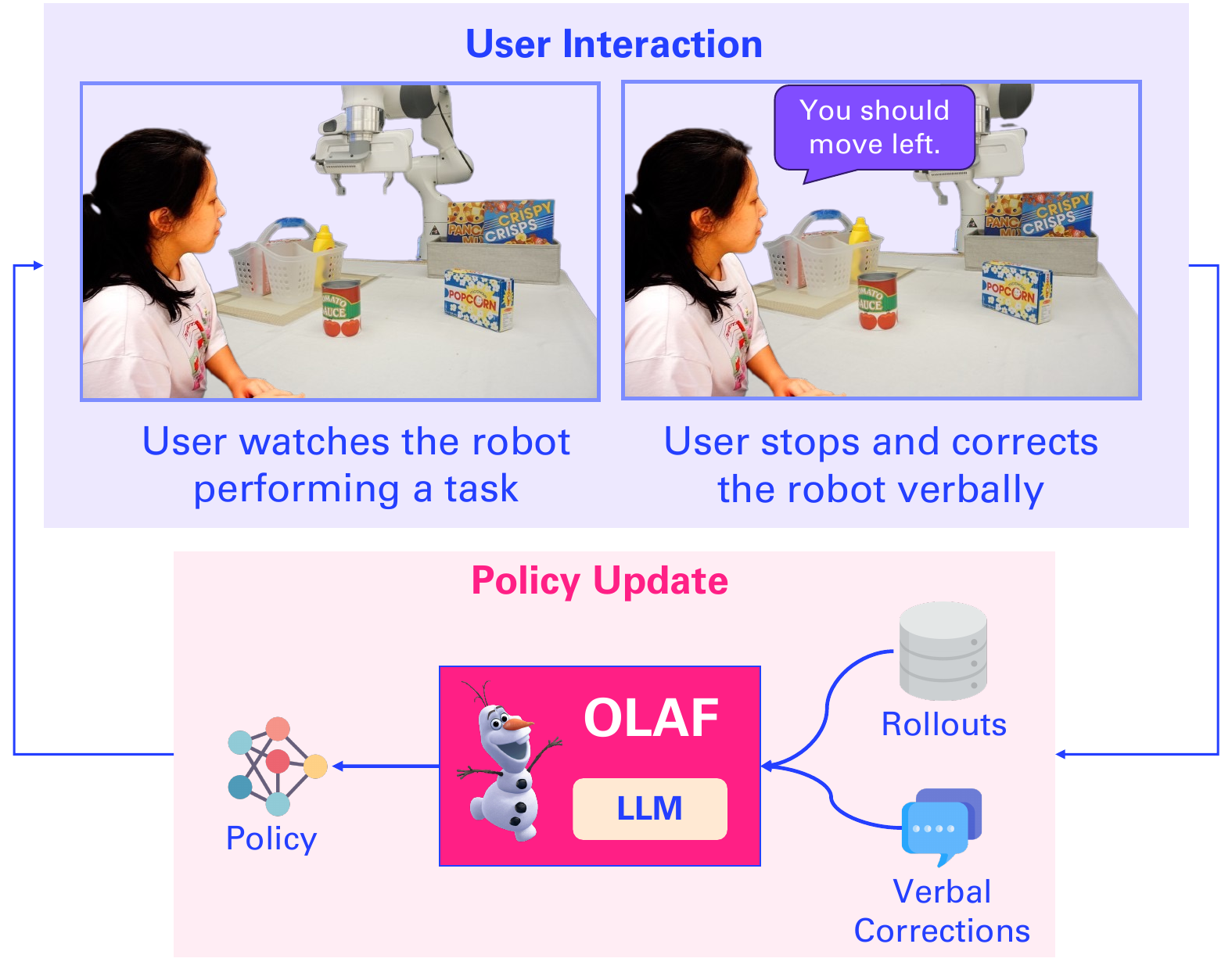}
    \caption{Teaching the robot through verbal correction with \methodname{}. \methodname{} is a LLM-based learning system designed for updating a robot's visuomotor neural-network-based policy using verbal corrections given by regular non-expert users. To train the robot, the user simply needs to watch to robot performing a task, stop the robot when the user thinks the robot is not able to finish the task, and then provide an instruction in natural language on how the robot can do better.
 }
    \label{fig:robot teaching}
    \vspace{-6mm}
\end{figure}

Language has long been recognized as an intuitive modality for people to provide feedback during robot learning across a variety of scenarios. Recent works have explored both learning language-conditioned policies to specify tasks to robots~\cite{yu2022using,singh2023progprompt,yu2023language, lynch2022interactive} and giving verbal corrections to a robot's course of action~\cite{Cui_2023,sharma2022correcting,yu2023language}. However, these approaches do not \emph{learn} from the verbal feedback -- the robot will repeat errors even after being corrected once. As such, they require perpetual correction \cite{lynch2022interactive, yu2023language} or shared control \cite{Cui_2023} from the user. \methodname{}, on the other hand, incorporates verbal corrections into the robot's neural policy using a learning algorithm, so that the robot avoids similar errors in the future.

\methodname{} builds on the rich literature on \Hcil{} (\hcil{})~\cite{osa2018algorithmic,zhang2018deep,kelly2019hg,mandlekar2020human,hoque2021thriftydagger,hoque2021lazydagger,momart_wong22a, liu2022robot} to train a robot policy. 
\hcil{} posits that robots need to learn and refine their behavior after deployment. 
Indeed, the deployment scenarios are so very diverse that collecting sufficient data to train a robot policy to \emph{perfectly} behave in all of the use cases is economically infeasible. 
With existing \hcil{} algorithms, the user can stop the robot's execution when it makes mistakes and demonstrate how to fix them. 
However, so far these techniques have been largely limited to scenarios where teleoperation or kinesthetic teaching is easy~\cite{zhang2018deep,mandlekar2020human,dass2022pato,momart_wong22a}. In everyday scenarios, this is rarely the case: these types of interventions require special equipment, skill, and/or physical strength in order to correct a robot's mistakes. 
Teaching a robot can be much easier and more intuitive if the robot can learn directly from natural language feedback.

\begin{figure*}[t]
    \centering
    \includegraphics[width=\linewidth]{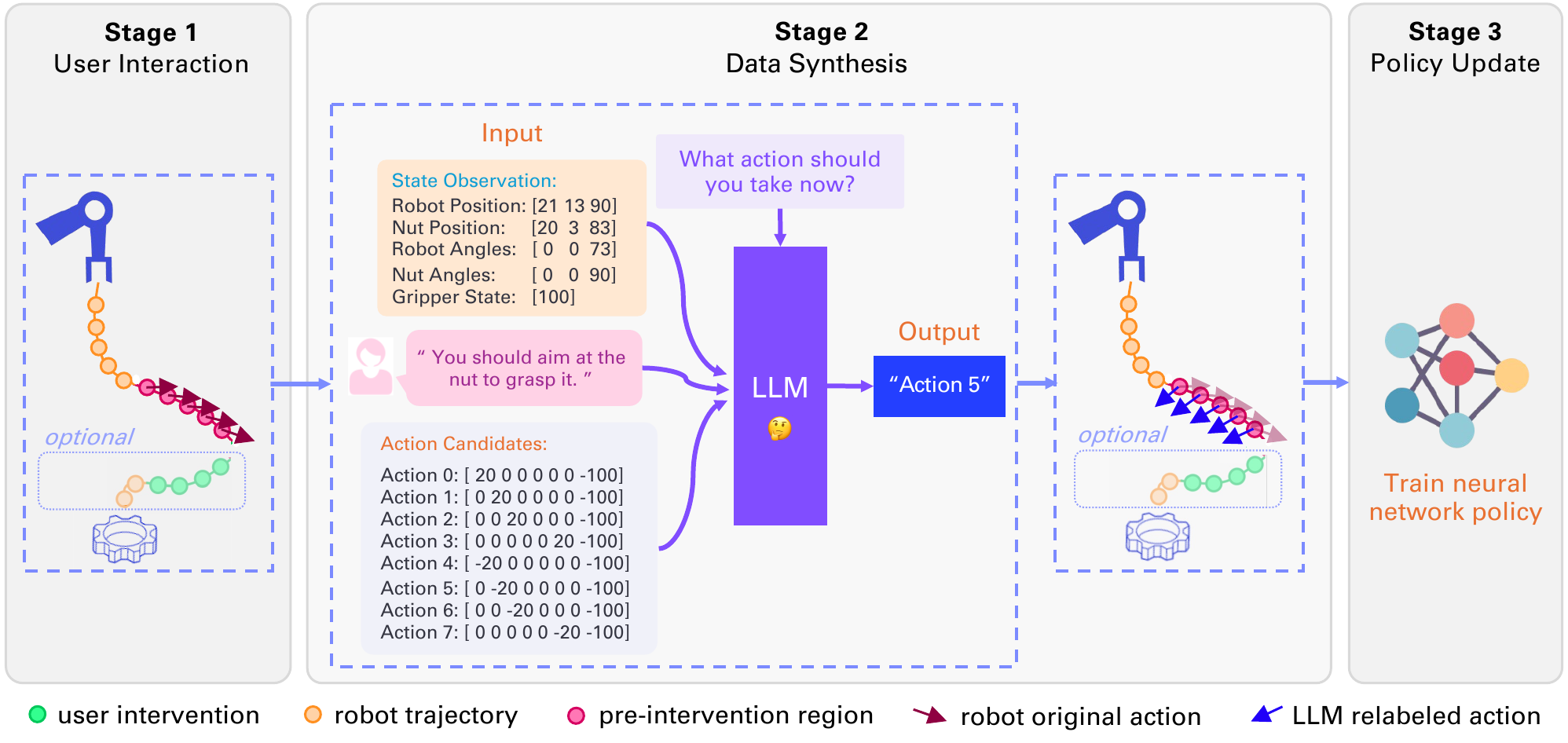}
    \caption{\methodname{} System. The \methodname{} pipeline consists of three steps: User Interaction, Data Synthesis, and Policy Update. 
    In User Interaction, it collects pairs of $\langle \mathit{robot\ trajectory}, \mathit{verbal\ correction} \rangle$ of trajectories stopped by the user. In Data Synthesis, it uses the LLM as a critic to select the action (from a pool of action candidates) that best matches the user's verbal correction and relabels the pre-intervention trajectory segments (in red).
    In Policy Update, it updates the policy by performing behavior cloning on the newly synthesized data and the previously collected data.
    }
    \label{fig:method}
\end{figure*}

\methodname{} is a \emph{learning} system that regular non-expert users can teach robots with natural language utterances. 
At a high-level, \methodname{} uses a setup analogous to \hcil{} but replaces physical intervention (e.g., provided via tele-operation) with verbal correction, as shown in \cref{fig:robot teaching}. In each episode, the robot attempts to complete a given task with its visuomotor policy, and the human user can stop the robot when necessary. After the robot stops, the user says what the robot should have done instead. 
As such, \methodname{} can be applied to scenarios where physical correction may not be feasible. %
\methodname{} still allows \emph{optional} physical corrections, as we demonstrate in \cref{sec:exp with intervention}.

The main technical novelty of our approach is an LLM-based \emph{action relabeling} strategy that improves the robot's policy from the user's verbal correction. 
When a user stops the robot, this is normally due the user noticing some wrong behavior of the robot. The wrong behavior is caused by the robot outputting actions that are potentially problematic, illustrated in \cref{fig:method} Stage 1 in {\textcolor{Burgundy1}{red}} ${\mathbf{\mathcolor{Burgundy1}\searrow}}$. Given a robot trajectory interrupted by the user, \methodname{} relabels the suboptimal actions \emph{in the trajectory segment leading up to the trajectory's termination} with good actions based on the user's verbal correction, illustrated in \cref{fig:method}, Stage 2 in {\textcolor{blue}{blue}} ${\mathbf{\mathcolor{blue}\swarrow}}$. \methodname{} employs an LLM to relabel the actions. 

A key highlight is that we rely on LLMs' ability to reason about \emph{non-verbal, numerical data} capturing the robot's and the world's spatial configurations and dynamics, not just on LLMs' commonsense reasoning. 
For each time step in the relabeled interval, \methodname{} presents the LLM with a set of candidate actions and uses the LLM as a critic, prompting it to select a better action from the candidate set  -- an action that is most consistent with the verbal feedback -- than the one the robot actually executed. Training the robot policy on the resulting relabeled actions allows the robot to learn counterfactual good actions that it \emph{should have} performed, and therefore perform better in the future.

We showcase the efficacy of \methodname{} both in simulation and on physical hardware, achieving on average $20.0\%$ improvement in policy success rate. Using verbal correction alone, without the user providing intervention physically, \methodname{} improves the success rate on average by $17.8\%$. In addition, the experimental results show that \methodname{} is compatible with existing \hcil{} methods and can improve their performance when user correction is available in addition to verbal correction. In summary, our main contributions are of follows:

\begin{itemize}[leftmargin=*]
    \item We introduce a \emph{learning} system that allows the human to \emph{improve} the robot's policy with verbal corrections. 
    \item We develop a novel action relabeling method where we employ an LLM to relabel the robot's erroneous actions with good actions according to human verbal corrections.
    \item We evaluate our method against baselines in simulation and on physical hardware and demonstrate its effectiveness at improving the robot policy after deployment.
\end{itemize}

\section{Learning from Verbal Correction} \label{sec:method}

\methodname{} (\cref{fig:method}) is a LLM-based learning system designed for updating a robot's visuomotor neural-network-based policy using verbal corrections. 
It trains the robot's policy in three steps: 
\begin{enumerate}[leftmargin=*]
    \item \textbf{(User Interaction)} It executes the robot's current policy in an attempt to finish the assigned task, while allowing the user to stop the robot anytime and give verbal correction (i.e., a natural-language instruction on how the robot could have better solved the problem). This is repeated several times to collect $\langle \mathit{robot\ trajectory}, \mathit{verbal\ correction} \rangle$ pairs.

    \item \textbf{(Data Synthesis)} Using this interaction data, \methodname{} synthesizes a training dataset to update the robot's policy. This is done by using LLM to incorporate the verbal correction to relabel the robot's executed action with the desired actions on the trajectory segments before the robot is stopped.

    \item \textbf{(Policy Update)} 
    It aggregates the newly synthesized data with the robot's existing data. Then it updates the robot's neural network policy by imitation learning on the aggregated data. 

\end{enumerate}

After training, \methodname{} internalizes the verbal corrections, translating them into changes to the weights of the robot's neural network policy. This helps the updated policy better complete the task and avoid the same mistakes that previously led the user to stop the robot.

This weight-updating approach of \methodname{} has several benefits over the common approach of storing the verbal feedback as prompts to control a LLM-based policy\footnote{A LLM-based policy or planner is a routine that queries an LLM API in every time step in execution.} \cite{huang2022inner,ahn2022can}.
First, \methodname{} can leverage the generalization ability of neural network policies. The verbal correction given by a user is typically related to how and where the robot makes mistakes. For instance, if the goal is on the left the robot, the verbal correction could be "Move to the left.", but following this instruction literally is not be meaningful in a new situation where the goal is on the right of the robot. As a result, it is important that robot in learning understands the context in which the verbal correction is given. 
Training the neural network policy on the synthesized data associates the robot's observations with the desired actions inferred from the verbal correction, which addresses the need of contextualization.
Second, \methodname{} only needs to query LLM offline as opposed to in real time, where the latter is needed for LLM-based policies. Offloading LLM queries to the offline training process allows the robot to run more smoothly with a higher control frequency, because the inference time of a typical visuomotor neural network policy is much faster and more conssitent than querying an LLM api ($<0.1s$ vs. $1$-$5s$). As a result, \methodname{} can achieve smoother robot motion than an LLM-as-policy approach.

Below, we describe the three steps of \methodname{} in details.

\subsection{User Interaction}\label{sec:example}

\begin{figure}[t]
    \centering
    \includegraphics[width=\linewidth]{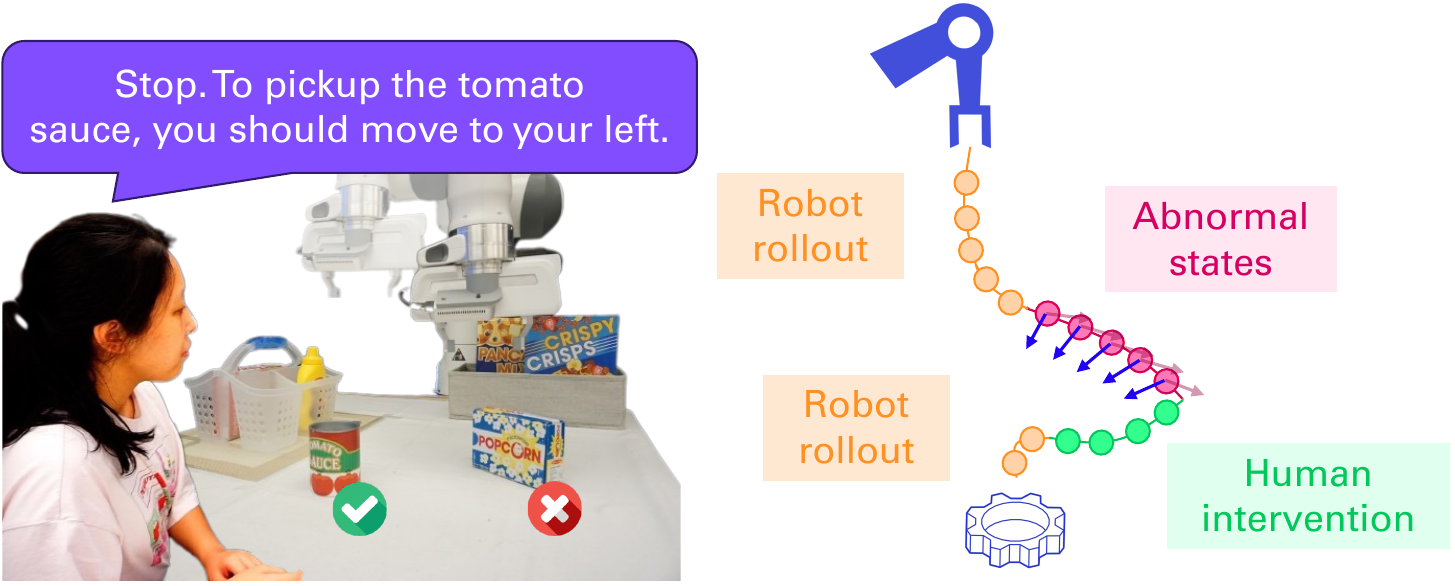}
    \caption{The left shows an example of a user interaction. The right shows how a trajectory is partitioned and the action in the pre-intervention region (abnormal state) is relabelled by the LLM. }
    \label{fig:example}
    \vspace{-5 mm}
\end{figure}

Given a pretrained visuomotor policy, \methodname{} collects training data to update the policy through interactions with a human user. Here we use an example to illustrate this process.
\cref{fig:example} depicts an use case of \methodname{} for finetuning a robot manipulator's policy. Here the robot is tasked to place the tomato sauce in the basket. The robot opens its gripper and moves forward. But instead of going to the tomato sauce, it goes to the right. Upon seeing this surprising behavior, the user stops the robot by pressing a stop button and says\footnote{In the experiments, our system takes the user's verbal correction through keyboard.} ``Stop. To pickup the tomato sauce, you should move to your left.'' 
If teleoperation or kinesthetic teaching is available, the user can optionally provide intervention (i.e. physical correction) to physically drive the robot to a better state solve the task and then finally release the control back to the robot.
This interaction episode can be repeated multiple times to collect a batch of interaction data.

\subsection{Data Synthesis: Action Relabeling Insight \label{sec:data_synth}}

\methodname{} uses LLM to relabel the robot's generated trajectory based on the user's verbal feedback and generate an imitation learning dataset from which the robot can learn to better solve the task.
But how do we achieve this effectively given that the user only provides the verbal correction once in a long trajectory? 
Our key insight is use the verbal correction to just label the desired actions of a short trajectory segment before the verbal correction is given.
The rationale for this design is the observation that there is often a delay between when the robot starts to make mistakes and when the user realizes the issue and stops the robot. 
This delay for an average user is around 1-2 seconds~\cite{liu2022robot}. We call this window the "pre-intervention region", which is where the error of robot is most likely to have happened and where the robot's actions are potentially wrong.
Before the pre-intervention region, the robot's trajectory reflects a nominal behavior to solve a task, which does not need correction. 
Based on this observation, in \methodname{}, we apply LLM to relabel the action in this "pre-intervention region", while keeping the starting robot trajectory as it is when synthesizing the data for policy update.

Specifically, let $o_t$ and $a_t$ denote the robot's observation and action at time $t$. Suppose the robot generated a trajectory $\xi = \{ o_0, a_0, \dots, o_T  \}$ and receives a verbal correction $v$, where  $T$ is when the robot is stopped by the user and given with the verbal correction $v$.  \methodname{} relabels the pre-intervention region $\{o_{T-K}, a_{T-K}, \dots, o_T, \hat{a}_T \}$ of size $K$ at the end of the trajectory with new actions suggested by the LLM based on the verbal correction $v$, which results in a modified trajectory 
$ \hat{\xi} = \{ o_0, a_0, \dots, a_{T-K-1}, o_{T-K}, \hat{a}_{T-K}, \dots, \hat{a}_{T-1}, o_T\}$, where $\hat{a}_t$ denotes the new action suggested by the LLM for time step $t$. 
This modified trajectory is than combined with existing training data (that the pretrained policy is based on) to update the policy through imitation learning.

For cases where intervention (i.e. physical correction) can be given by the user in addition to verbal correction, \methodname{} would include those into the training data. In this case, the relabeled data would take the form of $\{ o_0, a_0, \dots, o_{T-K}, \hat{a}_{T-K}, \dots, o_T, \tilde{a}_T, \dots \allowbreak \dots, o_{T+I}, a_{T+I}, \dots, a_{H-1}, o_H \}$, where $\tilde{a}_t$ denotes the intervention given by the user at time $t$, $I$ is the duration of the intervention, and $H$ is the length of the full trajectory. That is, the recorded trajectory would be a sequence of 
\begin{enumerate*}
    \item the robot's initial trajectory, where the mistakes have not happened;
    \item the pre-intervention region, which covers the mistakes;
    \item the user correction
    \item the robot's terminal trajectory after the user corrects the robot and releases back the control. 
\end{enumerate*}
The decomposition can be visualized in \cref{fig:example}.

\methodname{} relabels the pre-intervention region, while prior \hcil{} algorithms~\cite{liu2022robot,kelly2019hg} choose to discard these potentially wrong actions and learn mainly from the physical user intervention data.
In comparison, the relabeled actions in the pre-intervention region provide the signal to avoid the mistakes that caused the user to stop the robot, whereas the user intervention only shows how the errors can be corrected after occurrence. 
Therefore, by mimicking the relabeled data generated based on the user's verbal correction (as well as the user intervention), \methodname{} can learn to directly avoid the previous mistakes and better solve the task.

\subsection{Data Synthesis: LLM as Critic for Action Relabeling}
\label{sec:method-data-synthesis}

\begin{figure}[t]
    \centering
    \begin{subfigure}{\linewidth}
         \centering
        \includegraphics[width=\linewidth]{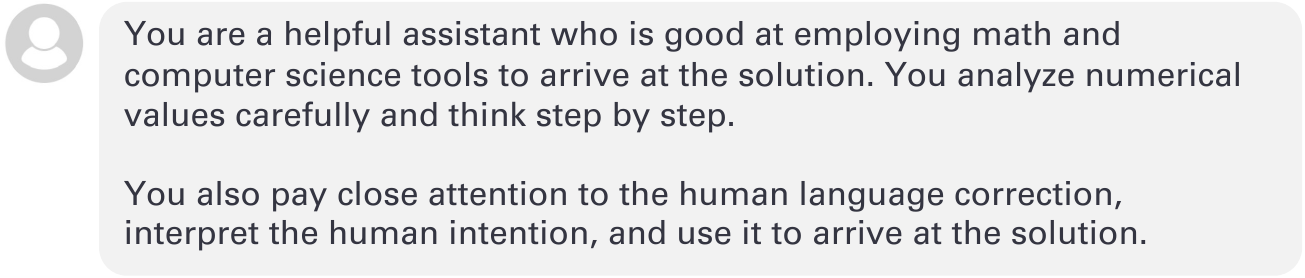}
         \caption{System prompt}
         \label{fig:system prompt} 
         \vspace{2mm}
     \end{subfigure}
    \begin{subfigure}{\linewidth}
         \centering
        \includegraphics[width=\linewidth]{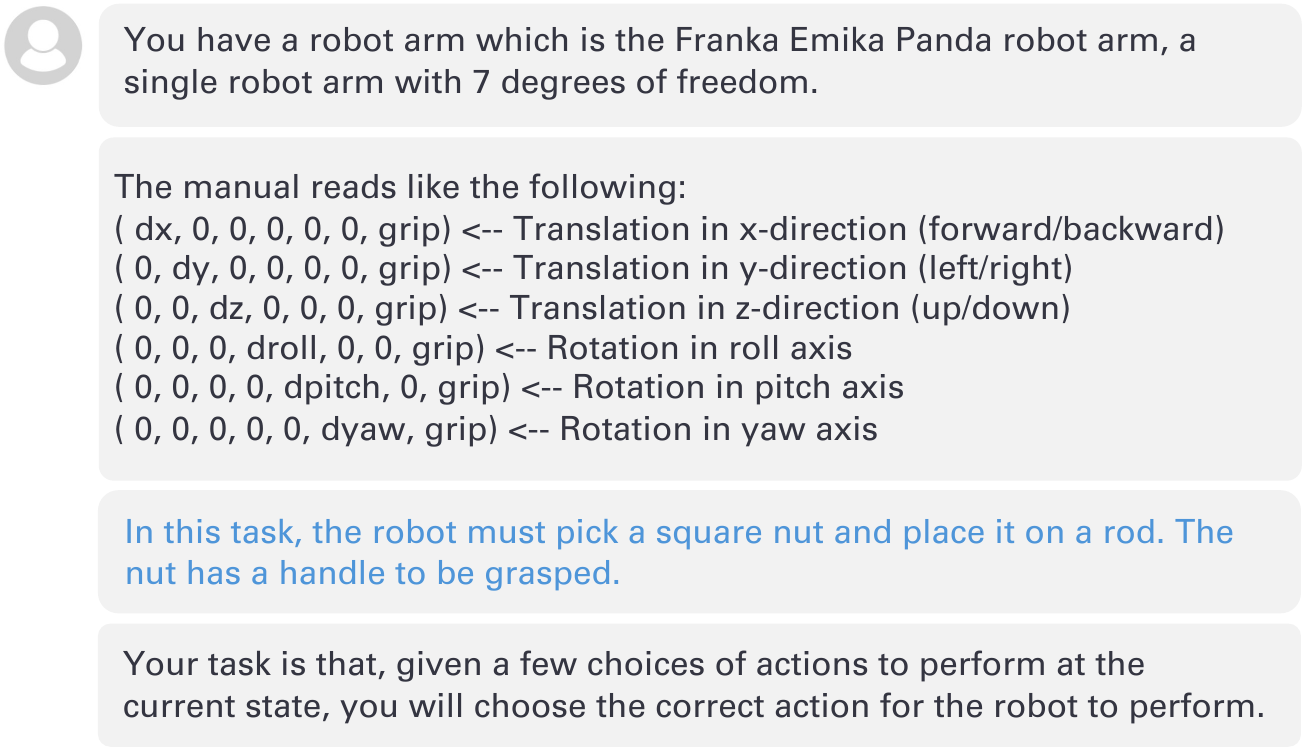}
         \caption{Context prompt}
         \label{fig:context prompt} 
         \vspace{2mm}
     \end{subfigure}
    \begin{subfigure}{\linewidth}
         \centering
          \includegraphics[width=\linewidth]{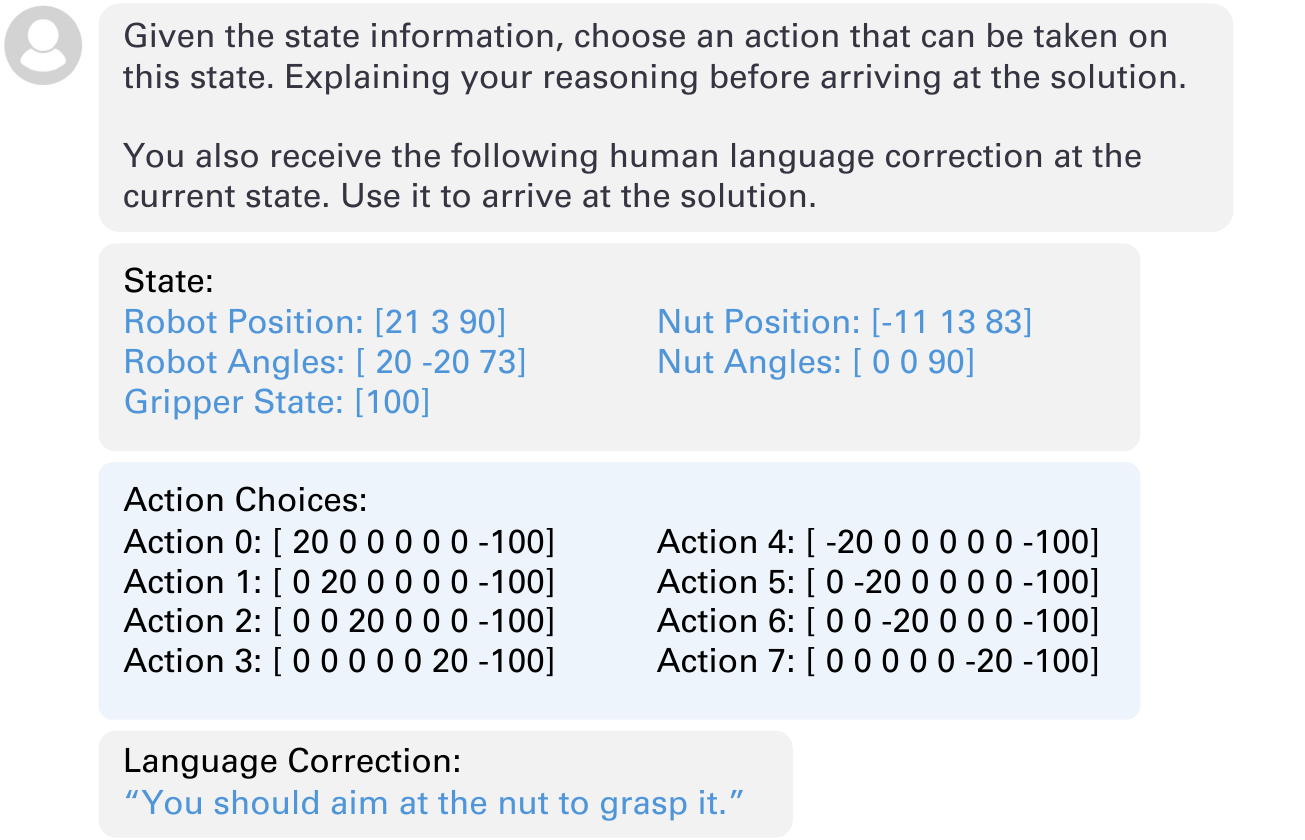}
         \caption{Action relabeling  prompt}
         \label{fig:action relabeling prompt}         
         \vspace{2mm}
     \end{subfigure}
    \caption{Prompts of an LLM as a critic for action relabeling. The system prompt specify system-level desired behavior, the context prompt describes the task level instruction, and the action relabeling prompt includes the trajectory-level information and the verbal correction.   
    The black denotes the template and the blue denotes user- or sensor-dependent information. We highlight the action proposal in blue background.}
    \label{fig:prompts}
    \vspace{-5 mm}
\end{figure}

\begin{figure*}[t]
    \centering
    \includegraphics[width=0.95\linewidth]{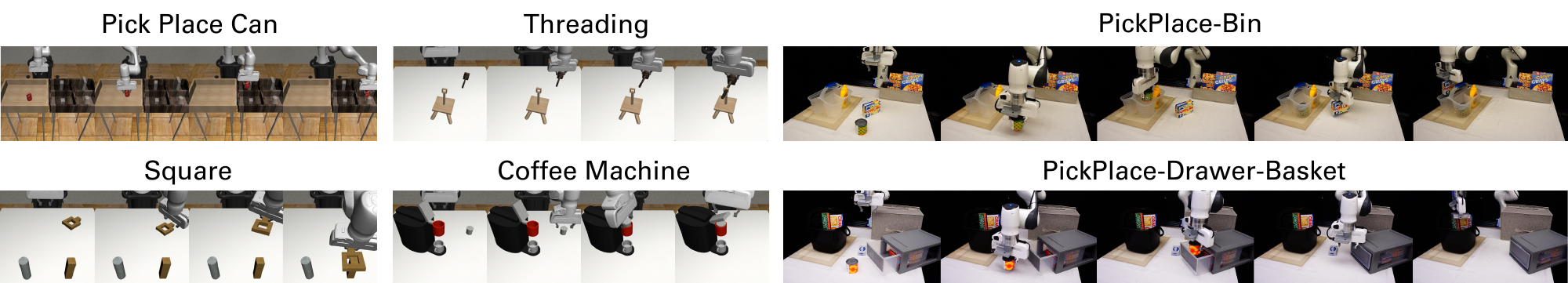}    
    \caption{We evaluate \methodname{} on four tasks in simulation and two tasks on real robot. The tasks in simulation are fine-grained manipulation tasks while the tasks on real robot are long-horizon, multi-staged tasks.}
    \label{fig:enter-label}
    \vspace{-5 mm}
\end{figure*}
\methodname{} relabels the pre-intervention region by using LLM as a \emph{critic} to select, from a set of action candidates, the action that best matches the verbal feedback. 
\cref{fig:prompts} shows the prompts used to query the critic LLM. It is composed of three parts: a system prompt, a context prompt and an action-relabeling prompt. The system prompt specifies system-level desired behavior for LLM, e.g. ``You are a helpful assistant with good analytical skills''. The context prompt provides the context for the task, including a basic manual of the robot (in black) and the task-dependent instruction (in blue). The action relabeling prompt reads state information from the robot's trajectory and the user's verbal feedback (in blue) and asks the LLM to select the best action from the candidates to achieve the task.
In the experiments, we consider action candidates that cover unit motion changes in all the degrees of freedom of the robot (the end-effector's position and the gripper state of a Franka Emika Panda robot).
Given a robot trajectory $\xi = \{o_0, a_1, \dots,  \dots, o_T  \}$, interrupted by the user at time $T$, we set the state in \cref{fig:prompts} as $o_T$ and then use the output of the LLM $\hat{a}$ to relabel all the actions in the pre-intervention time interval $[T-K, T-1]$. That is, we have the relabelled trajectory $\hat{\xi} = \{ o_0, a_0, \dots, a_{T-K-1}, o_{T-K}, \hat{a}_{T-K}, \dots, o_T,  \}$ and $\hat{a}_t = \hat{a}$ for all $t \in [T-K, T-1]$. 
We apply this procedure for each trajectory collected from the user interaction phase individually and create the synthetic data that will be used to update the robot's policy.

There are several ways to generate the list of action candidates for the LLM to choose from. We find that the following simple action proposal method works reasonably well: we generate a list of ``one-dimensional" actions, each of which makes a change to a single dimension in position, orientation, or gripper state, and let the LLM choose from this list (see \cref{fig:action relabeling prompt}). Rather than using these ``one-dimensional" actions directly, we apply them as delta-actions on top of the original policy's actions. We scale the actions to be integers, which are more intuitive for an LLM to interpret and are also short in terms of LLM token length. Ablation results of different action proposal methods can be found in \cref{sec:analysis-ablation}; more details on the LLM workflow design and hyperparameters can be found in Appendix \ref{appendix:llm_workflow}.

Notice that \methodname{} queries the LLM only once per verbal correction. 
We found that issuing one query and then applying the results to all time steps in the \emph{pre}-intervention period achieves similar performance to issuing a separate query for each individual time step, but the former is significantly more cost effective (about $15\times$ fewer LLM calls). We present more analysis in \cref{sec:analysis-ablation}.

\subsection{Policy Update}

To update the policy, \methodname{} combines the relabeled data and the dataset used to pretrain the robot's initial policy together as an aggregated dataset $\hat{D}_+$. Then it fits the policy on the aggregated dataset (which consists of sequences of observations and actions) using behavior cloning: 
\begin{equation}
    \max_\pi \mathbb{E}_{ h,a \sim \hat{D}_+} [ \log \pi(a|h) ]
\end{equation}
where $h$ denotes the history of observations and action preceding the action $a$, and $\mathbb{E}_{ h,a \sim \hat{D}_+}$ denotes the expectation over the data distribution of $\hat{D}_+$.  In practice, this can be done using stochastic gradient descent and minibatch sampling.

In general, we can run \methodname{} in multiple rounds (of user interaction, data synthesis, and policy update), where the latest policy is used to collect new user-interaction data. 
Therefore, we can view \methodname{} as an extension of the interactive imitation learning algorithm, DAgger~\cite{ross2010reduction}. 
The main difference is that \methodname{} uses LLM (conditioned on the user's verbal feedback) as the expert policy, as opposed to using the user as the expert directly. Past studies have found that human users are incapable of providing high-quality feedback to relabel robot trajectories due to the missing sensorimotor feedback. By using LLM and having the user provide verbal correction instead, we circumvent this issue without introducing tele-operation or kinesthesis teaching, which require additional setups.
In the experiments, we uses only one pass of data collection and policy update.

\begin{table*}[ht]
    \centering
    \caption{\textbf{Experimental results of learning with \emph{only} verbal correction.} \methodname{} consistently improves the performance over the baseline of self-imitation. \methodname{} works the best when the feedback is detailed (long feedback) compared with high-level verbal correction (short-feedback). Running \methodname{} but querying the LLM critic to select the action without the user instruction (at the end of \cref{fig:prompts}) performs worse than \methodname{} conditioned on the user feedback. We report the mean success rate and the standard deviation of three random seeds.}
    \begin{tabular}{l|c|c|c|c}  
        \toprule
        \textbf{Method} & \texttt{Pick Place Can} & \texttt{Threading} & \texttt{Square} & \texttt{Coffee Machine}  \\        
        \midrule
        BC (self-imitation) & $73.6 \pm 4.5$  & $53.3 \pm 3.8$ & $41.0 \pm 2.2$ & $16.0 \pm 2.8$ \\
        \methodname{} (no feedback)  & $81.3 \pm 3.4$ & $51.0 \pm 0.8$ & $44.0 \pm 7.2$ & $45.3 \pm 2.5$  \\
        \methodname{} (short feedback)  & $82.6 \pm 1.9$ & $51.0 \pm 1.4$ & $48.7 \pm 1.2$ & $47.3 \pm 1.9$  \\
        \methodname{} (long feedback) & $\textbf{84.6} \pm 3.1$ & $\textbf{60.5} \pm 2.5$ & $\textbf{59.0} \pm 7.1$ & $\textbf{51.0} \pm 1.4$ \\
        \bottomrule
    \end{tabular}
    \label{tab:verbal-only}
\end{table*}

\begin{table*}[ht]
    \centering
    \caption{\textbf{Experimental results of learning with \emph{both} verbal correction and intervention.} 
    The upper part compares algorithms that do not introduce data weighting. Comparing Imitation and \methodname{} shows the benefit of also learning from the relabelled data based on verbal correction.
    The bottom part compares recent \hcil{} algorithms that use data weighting and the hybrid version of \methodname{} and Sirius~\cite{liu2022robot} (i.e., \methodname{} using the weighting scheme of Sirius). Generally, \methodname{}+Sirius performs the best, even slightly better the state-of-the-art Sirius \hcil{} algorithm. These results show again the effectiveness of \methodname{} incorporating verbal correction.
    We report the mean success rate and the standard deviation of three random seeds.}    
    \begin{tabular}{l|c|c|c|c} 
        \toprule
        \textbf{Method} & \texttt{Pick Place Can} & \texttt{Threading} & \texttt{Square} & \texttt{Coffee Machine}  \\  
        \midrule
        BC (self-imitation) & $95.3 \pm 3.1$ & $84.4 \pm 6.2$ & $66.6 \pm 2.3$ & $78.8\pm 1.7$ \\
        \methodname{}  & $\textbf{97.0} \pm 1.4$ & $\textbf{86.5} \pm 1.8$  & $\textbf{76.3} \pm 3.5$ & $ \textbf{84.0} \pm 2.8$ \\
        \midrule
        HG-DAgger~\cite{kelly2019hg}  & $\textbf{95.4} \pm 1.4$ & $75.0 \pm 2.5$ & $72.1 \pm 4.0$ & $75.4 \pm 3.8$ \\
        IWR~\cite{mandlekar2020human}  & $94.6 \pm 2.0$  & $86.7 \pm 2.9$ & $78.8 \pm 2.7$ & $85.8 \pm 2.0$ \\
        Sirius~\cite{liu2022robot}  & $95.0 \pm 3.8$ & $87.8 \pm 3.6$ & $80.4 \pm 2.0$ & $86.3 \pm 3.0$ \\
        \methodname{} + Sirius & $95.0 \pm 1.3$ & $\textbf{88.3} \pm 1.4$  & $\textbf{82.1} \pm 0.7$ & $\textbf{87.9} \pm 0.7$ \\
        \bottomrule
    \end{tabular}
    \label{tab:with-intervention}
    \vspace{-0.5cm}
\end{table*}

\section{Related Work}

\subsection{Human-in-the-loop Imitation Learning}

Data Aggregation (DAgger)~\cite{ross2011reduction} is a canonical interactive imitation learning technique to update policies in deployment. 
DAgger first executes the pretrained policy in an environment, relabels the generated trajectories in hindsight with an expert policy's actions, and retrains the learner policy on the relabeled data (and the original data).
However, humans struggle to play the role of the expert in DAgger because 
human sensorimotor control relies on receiving timely feedback, making hindsight relabeling difficult in practice and to potentially degrade robot learning performance~\cite{laskey2017comparing}. 
\hcil{} algorithms~\cite{kelly2019hg,spencer2020learning,mandlekar2020human,liu2022robot} address this issue through \emph{intervention}, which gives the human user full control to correct the robot (e.g., via tele-operation) at the moment when the user thinks the robot is starting to do something wrong. Compared to hindsight relabeling in DAgger, intervening and correcting is more intuitive for humans. 
HG-DAgger~\cite{kelly2019hg}, based on DAgger, learns from 
interventions, but it only updates the learner policy on the intervention. IWR~\cite{mandlekar2020human} trains the policy also on the robot's own trajectories before the intervention (with lower importance weights), which improves the stability and performance over HG-DAgger. Sirius~\cite{liu2022robot} further 
removes a small pre-intervention trajectory segment because human users have a minimum reaction time before they can intervene on the robot. 
EIL~\cite{spencer2020learning} follows a similar idea but trains value functions to update the policy, rather than directly imitating interventions.

Our approach \methodname{} can also use intervention, so it can be viewed as a \hcil{} method. However, unlike these previous approaches, \methodname{} can learn from verbal correction as well. 
As a result, providing interventions physically (though helpful) is optional with \methodname{}, which is good for scenarios, e.g., where tele-operation is infeasible to set up.
Moreover, 
\methodname{} learns to pre-empt mistakes instead of only correcting them. This is accomplished by using the verbal feedback to relabel a trajectory segment \emph{before} the user intervention.

\subsection{Instruction Following / Language-Conditioned Policies}

Despite being more ambiguous than demonstrations, natural language is perhaps the most intuitive modality to instruct a robot. Many recent works have designed instruction-following robots that allow the user to control the robot via natural language commands.
One approach is to pre-train a language-conditioned policy using (self-)supervised learning from pairs of demonstrations and task instructions \cite{yu2022using,sharma2022correcting}. 
Another approach is to leverage an LLM to interpret and reason about verbal instructions: 
\cite{sundaresan2023kite} uses LLM to parse instruction into way-points;
 \cite{liang2023code,vemprala2023chatgpt,singh2023progprompt} use LLM to generate program codes to control the robot, whereas \cite{huang2022inner} and \cite{ahn2022can} use LLM to decompose the problems into small steps. \cite{yu2023language} uses LLM to generate rewards for online planning, and \cite{ren2023robots} further uses LLM to ask for clarification when the instruction is vague.

However, none of these methods learn from instructions to improve the pre-trained policy. 
In other words, these systems function more or less as language-based tele-operation. While verbal correction studied in this paper can be viewed as a form of verbal instruction, \methodname{} significantly differs from the aforementioned methods in that it uses LLM to \emph{update} neural visuomotor policies.
Our approach showcases that an LLM's non-verbal pattern recognition ability can be used in conjunction with its reasoning ability (which interprets the verbal feedback) to train neural networks. 
As a result, a robot trained by \methodname{} would internalize the corrected motor skill and can complete the task alone without requiring constant human supervision.
    
\subsection{LLM for Labeling and Finetuning ML Policies}

LLMs have recently emerged as generalist agents for many tasks that can be expressed in language. For instance, they have been used to annotate supervised learning datasets~\cite{bansal2023large,he2023annollm} using their in-context learning capabilities. \methodname{} also uses LLM to annotate robot trajectories, but leans on their physics and common-sense reasoning rather than in-context learning.
In parallel, the high quality of LLM annotations has birthed a very prolific area of LLM distillation, where labeled datasets created by querying LLMs are used to train smaller language models~\cite{chang2023learning,hsieh2023distilling,lee2023rlaif}. \methodname{} differs from them in that the learner model is a neural visuomotor policy (implemented by BC-transformer) rather than a language model.

\section{Experiments}
\label{experiments}

Our empirical study of \methodname{}'s effectiveness focuses on two questions: \textbf{(1)} Can \methodname{} effectively learn to improve a pretrained visuomotor policy by using \emph{only} verbal feedback? \textbf{(2)} Can learning from verbal correction with \methodname{} still be useful when human physical intervention is available? We conduct experiments on both simulation tasks that involve fine-grained manipulation and real-robot tasks that are long-horizon and multi-stage.

\subsection{Setup}
\label{sec:setup}

Our interactive learning experiment setup goes as following. For each task, we collect $M$ human demonstrations via teleoperation and pretrain a visualmotor policy with multimodal inputs: 1 workspace camera image, 1 eye-in-hand (wrist) camera image, and robot proprioceptive state.
In each experiment, we use the pretrained policy for the task, while a human user monitors the robot. During the robot's execution, if the user thinks that the robot cannot finish the task, the user stop the robot and provide a verbal correction (through keyboard) to describe what or how the robot should have done instead before it was intervened. The user can optionally provide physical intervention by taking over control from a teleoperation device like Spacemouse, then release the control back to the robot. 
In the experiments, the verbal corrections are provided via a keyboard, but in real-world deployments we envision using voice recognition systems such as Whisper~\cite{radford2022whisper} for this purpose.
We repeat this process to collect $N$ trajectories with verbal corrections. The interventions are provided by a PhD student with robotics experience. The verbal corrections are provided by a recruited senior CS undergraduate student who has no robotics experience. The final policy is trained on the aggregated dataset of $M+N$ trajectories. 
We use GPT-4 as the LLM with temperature = 0.5. 
We use a transformer policy (resnet-18 encoder, spatial softmax, GMM head) with history length 10 (about 19M parameters in total). More hyperperameter details on policy and tasks can be found in \cref{appendix:policy_imp} in Appendix. \\

\noindent
\textbf{Simulation Experiments.} For simulation experiments, we evaluate \methodname{} on four manipulation tasks (\texttt{Pick Place Can}, \texttt{Threading}, \texttt{Square}, \texttt{Coffee Machine} in \cref{fig:enter-label}) based on robomimic~\cite{mandlekar2021matters}, a simulated robotic manipulation benchmark. In each task, the robot arm is a simulated Franka Emika Panda robot with 7 degrees of freedom (end-effector position, orientation and gripper state). We use $M = 50$ and $N = 100$. We use ground truth object state from the simulator for LLM prompting, and use image observation for policy learning. We train 3 seed for each method for 1000 epochs, and perform 50 trials of task execution every 100 epoches and compare the averaged best success rate across seeds for each method.\\

\noindent
\textbf{Real Robot Experiments.} 
For real robot experiments, we evaluate \methodname{} on two long-horizon tasks on a physical Franka Emika Panda arm: (1) pick up the pea can and popcorn and place them into the bin in sequence (\texttt{PickPlace-Bin}); and (2) pick up the peach can and place it in the drawer, and pick up the chocolate box and place it into the basket (\texttt{PickPlace-Drawer-Basket}) (see \cref{fig:enter-label}). The arm is with 5 degrees of freedom (end-effector position, yaw orientation and gripper state). We use $M = 40$ and $N = 80$. We obtain object pose information using 6D object pose estimator DOPE~\cite{tremblay2018deep} for LLM prompting, and use image observation for policy learning similar to simulation experiments. We use one seed, run each training method for 1000 epochs,  perform 34 task execution trials, and compare the methods' success rates on the last epoch's checkpoint.

\subsection{Learning from Verbal Correction Only}

In this experiment, we aim to study whether verbal correction provides useful information and can help the robot refine its policy's performance. In all the experiments here, we consider the setting with \emph{only} verbal correction. We only use the trajectory segments \emph{before} the human stops the robot, and ignore all trajectory segments of human intervention. The policy learns from the aggregated dataset of (1) the robot's rollout trajectory without intervention and (2) the initial expert demonstrations. We consider a self-imitation baseline, which updates the policy by imitating from the same dataset but without action relabeling.

We compare \methodname{} with long and short feedback. A long feedback refers to the user giving detailed directional instructions  (e.g., ``Move toward to the left to grasp the cup.''). A short feedback on the other hand refers to the user giving high-level, object-centric instruction (e.g., ``Move closer to the cup.'').
We consider these two kinds of feedback to test the limits of our LLM action relabeling. Using the short feedback is harder, as LLM additionally has to do some physics and commonsense reasoning. 
For ablation purposes, we also consider a baseline that runs \methodname{} without verbal feedback. We remove the part of the prompt about verbal correction in \cref{fig:prompts} and ask the LLM to select from among the action candidates directly based on the task and the current state.

The experimental results are summarized in \cref{tab:verbal-only}.
\methodname{} consistently improves the performance over the baseline of self-imitation. It works the best with long feedback. As expected, the no-feedback baseline is the weakest version of \methodname{} (though still better than self-imitation). This shows the importance of verbal feedback for the LLM (GPT4) to provide action suggestions.

\begin{table}[t]
    \centering
    \caption{\textbf{Experiment results on physical hardware.} We evaluated success rates on learning with both verbal correction and intervention. \methodname{} outperform BC baseline on both tasks, showing that the ability to correct mistakes so as to avoid iterating wrong behaviors is critical.}
    \begin{tabular}{l|c|c}  
        \toprule
        \textbf{Method} & \texttt{PickPlace-Bin} & \texttt{PickPlace-Drawer-Basket} \\        
        \midrule
        BC & $35.3$ & $52.9$ \\
        \methodname{} & \textbf{73.5} & \textbf{70.6} \\
        \bottomrule
    \end{tabular}
    \label{table:real-robot-results}
\end{table}

\subsection{Learning from Verbal Correction and Intervention}
\label{sec:exp with intervention}

In this experiment, we study whether verbal correction is still useful when human intervention, i.e., physical correction \emph{in addition to} stopping the robot, can be provided. As before, we consider a baseline, BC (self-imitation), which imitates the entire trajectory, including the parts generated by the robot and by human intervention. The experimental results are summarized in \cref{tab:with-intervention}. \methodname{} generally outperforms the BC baseline, showing the benefit of learning from verbal correction. We note that even when the robot can learn from data that corrects  its behavior after it commits a mistake, \methodname{} with action relabeling still outperforms the BC version without it. This highlights the importance of overriding erroneous actions to learn the accurate actions that prevent the mistakes from happening, rather than merely recovering from the mistakes.

Next, we consider three state-of-the-art \hcil{} algorithms (HG-DAagger~\cite{kelly2019hg}, IWR~\cite{mandlekar2020human}, and Sirius~\cite{liu2022robot}) which do not use verbal corrections, relying just on interventions. These algorithms mainly differ in how they weigh different parts of the trajectory in doing imitation learning. 
We combine \methodname{} with Sirius by incorporating the weighting scheme in Sirius~\cite{liu2022robot}, which up-weights interventions and down-weights  pre-interventions. The hybrid version performs on par with and slightly better than Sirius, showing that \methodname{} is compatible with state-of-the-art \hcil{} algorithms when human intervention is available. We also note that it is not a significant improvement because of how down-weighting pre-intervention samples minimizes the effect of action relabeling.

\begin{table}[t]
    \centering
    \caption{\textbf{A comparison of different action proposal methods on \texttt{Square}.} Querying LLM for ``one-dimensional'' delta actions (Onedim + Original) achieves better performance.}
    \begin{tabular}{l|c|c}  
        \toprule
        \textbf{Method} & Only Verbal & Verbal \& Intervention \\
        \midrule
        BC baseline & $41.0 \pm 2.2$ & $66.6 \pm 2.3$ \\
        \methodname{}: LLM Gives Actions & $49.3 \pm 1.2$ & $65.2 \pm 4.2$ \\
        \methodname{}: LLM Edits Actions & $52.0 \pm 2.8$ & $70.7 \pm 4.6$ \\
        \methodname{}: Onedim Actions & $54.0 \pm 3.5$ & $62.0 \pm 3.5$\\
        \textbf{\methodname{}: Onedim + Original} & $\textbf{59.0} \pm 7.1 $ & $\textbf{76.3} \pm 3.5$\\
        \bottomrule
    \end{tabular}
    \label{table:action-proposal-comp}
\end{table}

\begin{table}[t]
    \centering
    \caption{\textbf{A comparison of Basic and Full relabeling on \texttt{Square}.} The Basic version achieves on-par performance with the Full version.}
    \begin{tabular}{l|c|c}  
        \toprule
        \textbf{Method} & Only Verbal & Verbal \& Intervention \\
        \midrule
        \methodname{}: Basic & $59.0 \pm 7.1$ & $76.3 \pm 3.5$ \\
        \methodname{}: Full & $60.7 \pm 6.1$ & $78.0 \pm 3.5$ \\
        \bottomrule
    \end{tabular}
    \label{table:Basic-Lavish-comp}
\end{table}

\subsection{Experiments on Physical Hardware}

 We perform the evaluation in the setting of learning with both verbal correction and human intervention, and use the long-language-feedback version of \methodname{}. As shown in \cref{table:real-robot-results}, \methodname{} with language relabeling outperforms BC on the original data. It is noteworthy that \methodname{'s} large performance gain is achieved by relabeling just a few timesteps before intervention -- a very small fraction of the training data. We hypothesize that this disproportionately positive effect is due to that \methodname{'s} behavior corrections prevents the learning updates from internalizing the erroneous action choices, and that these real robot experiments requires a longer manipulation sequence. 
The BC baseline does not learn to override the robot mistakes themselves, but rather to correct them afterwards. For example, we found that if the robot accidentally drops the object, the BC policy would preserve the wrong gripper release action, and commit the same mistake again, which might not be recovered. On the contrary, \methodname{} uses the verbal correction to fix the error before it happens, allowing the robot to proceed to the next stage of manipulation.
On the other hand, the BC baseline does not learn to override the robot mistakes themselves, but rather to correct them afterwards; the robot does not avoid learning from the mistakes. For example, if the robot accidentally drops the object, it can preserve the wrong gripper release action, and commit the same mistake again which might not be recovered.

\subsection{Analysis and Ablation Studies}
\label{sec:analysis-ablation}

\noindent
\textbf{Comparison of different action proposal methods.} As discussed in \cref{sec:data_synth}, we query the LLM by asking for a ``one-dimensional" (\emph{onedim}) delta action, and add the delta action to the original action as the final relabeled action used for neural network training. In this section, we discuss alternative designs for obtaining relabeled actions:
\begin{enumerate}[leftmargin=*]
    \item LLM Gives Actions: Use the LLM as an actor and directly ask it to produce a 7D action.
    \item LLM Edits Actions: Use the LLM as an actor, tell it the agent's original action, and ask it to modify the action.
    \item Onedim Actions: Use the LLM as a critic, let it choose from a set of one-dimensional actions (e.g. moving in the positive x-direction)
    for moving in the positive x-direction), and its choice this as the final action. 
    \item Onedim Actions + Original (Ours): Query the LLM similar to Onedim Actions, but instead use the LLM's choice as a delta action for adding to the original action.
\end{enumerate}

We present the policy performance of each method in \cref{table:action-proposal-comp}. We find that most action proposal methods allow policy learning to outperform the BC baseline, especially for the Only Verbal version. However, using Onedim + Original yields the largest improvement over BC. One potential explanation is that using the LLM as a critic is generally more effective than using it as an actor directly, because a finite set of action choices provides structure and reduces hallucinations. Also, simple one-dimensional actions may be more intuitive for the LLM to understand than actions that attempt to modify multiple state dimensions at a time. Finally, human corrections are by definition modifications w.r.t. the existing policy actions, and the Onedim + Original method reflects this intuition closer than Onedim Actions, which may explain why Onedim + Original performs better. 
\\

\noindent
\textbf{Trade-off between LLM query quota and relabeling accuracy.} As mentioned in \cref{sec:method-data-synthesis}, we query the LLM only once for each human verbal correction to make running the system more cost-effective. Namely, we query for the last time step in each pre-intervention period, obtain an action choice from the LLM, and use that action for \emph{all} time steps in the pre-intervention region. A more elaborate -- but much more expensive -- version would be to issue a separate LLM query for each time step in the pre-intervention period so as to obtain separate relabeled actions. We call ours and the elaborate version the ``\methodname{}: Basic'' and ``\methodname{}: Full'' version, respectively. We compare the two methods in \cref{table:Basic-Lavish-comp} for the \texttt{Square} task, and discover that our version (Basic) achieves similar performance to the Full version while being much more cost-effective. At the same time, we recognize that our comparison was done under the assumption that the same action can apply to all pre-intervention time steps, which might not always be true and might generate compromised action relabeling results.

\section{Conclusion and Limitations}

We introduced \methodname{} -- the first learning system that can update visuomotor neural network policy using verbal correction from regular non-expert users. \methodname{} uses an LLM to translate verbal corrections into low-level action labels to synthesize a dataset for updating the policy. Experimentally, we show that \methodname{} is effective in leveraging verbal correction to improve policy performance, achieving on average $20.0\%$ improvement from BC baseline across simulation and physical hardware.
The current design of \methodname{} has several limitations. First, although we use \methodname{} to train \emph{visuomotor} transformer-based policies, the LLM requires \emph{textualized} state estimation to relabel actions. In addition, we need to hand-craft the task-dependent prompt properly for the LLM to understand the state information. While these limitations are common in existing LLM-related applications, we hope that they can be relaxed in the future.
\section*{ACKNOWLEDGMENT}
We thank Yue Wu for the helpful feedback and insightful discussions. We thank Ricky Loynd for the support and assistance on LLM infrastructures. We thank Ajay Mandlekar for sharing well-designed simulation task environments. This work was partially done during Huihan Liu's internship at Microsoft Research. We acknowledge the support of National Science Foundation (2145283, 2318065), the Office of Naval Research (N00014-22-1-2204) and Amazon.

\printbibliography

\clearpage
\onecolumn

\section*{Appendix}

\section{LLM Workflow Implementations}
\label{appendix:llm_workflow}

\begin{figure*}[ht]
    \centering
    \includegraphics[width=0.9\linewidth]{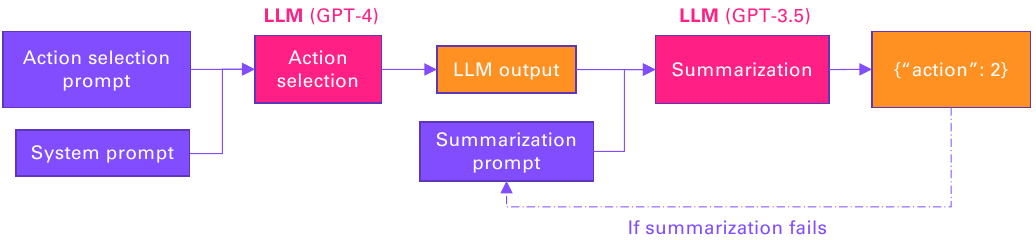}
    \caption{\textbf{\methodname{}'s Workflow for Querying LLM.} The \methodname{} LLM workflow consists of two steps: Action Selection and Summarization.  
    In Action Selection, a LLM inputs the action selection prompt and system prompt and outputs its CoT reasoning process along with its action choice. In Summarization, another LLM query is made to summarize the LLM reasoning output into a single action in json format. We use GPT-4 for Action Selection and GPT-3.5 for Summarization.
    }
    \label{fig:llm_workflow}
\end{figure*}

We present the pipeline of querying LLM for action relabeling previously discussed in Section \ref{sec:data_synth}. As shown in Figure \ref{fig:llm_workflow}, the \methodname{} LLM workflow consists of two steps: Action Selection and Summarization. In Action Selection, a LLM inputs the action selection prompt and system prompt, and outputs its unstructured response. We first make a query on gripper action selection, and LLM will return the correct gripper action (open or close). We then make a second query on the 6D action (x, y, z, roll, pitch, yaw) with the correct gripper action appended to it. 

The response contains its chain-of-thought reasoning process for the final action choice, which we need to parse and summarize into some structured form. Therefore, we introduce a second step of summarizing the unstructured LLM output into json format, which we found is a good template for GPT to produce structured response. In Summarization, another LLM query is made to summarize the LLM reasoning output into a single action in json format. This process can potentially fail, for example, failing to produce a single action, producing the wrong object type, which causes error in obtaining the original action later on. If this fails, we add a corrective feedback to the conversation and query at most 3 times until the response is correct (e.g. ``This is incorrect format. You should return the answer as single JSON object and the value should be a single number! Please try again.''). We use GPT-4 for Action Selection and GPT-3.5 for Summarization.

We use temperature = $0.5$ for all LLM queries, and relabel $T = 15$ timesteps in the pre-intervention region that correspond to the average human reaction time of 2 second \cite{liu2022robot}. We use $0.2$ for the 6D action scale, which is around the average value of robot actions.
\clearpage

\section{Examples of Prompts and LLM Outputs}

\subsection{Prompt for Gripper State Selection}

\begin{quotebox}[H]
    \begin{myquote}

You are a helpful assistant who pay close attention to the human language correction, interpret the human intention, and use it to arrive at the solution.

\vspace{4mm}

You have a robot arm which is the Franka Emika Panda robot arm, a single robot arm with 7 degrees of freedom.
The robot a parallel-jaw gripper equipped with two small finger pads, that comes shipped with the robot arm.
The robot comes with a controller that takes in actions. 

The expected action space of the OSC\_POSE controller (without a gripper) is \texttt{(dx, dy, dz, droll, dpitch, dyaw)}. 

\vspace{4mm}

The manual reads like the following: 

\texttt{( dx,  0,  0,  0,  0,  0, grip)     <-- Translation in x-direction (forward/backward) }     

\texttt{(  0, dy,  0,  0,  0,  0, grip)     <-- Translation in y-direction (left/right) }

\texttt{(  0,  0, dz,  0,  0,  0, grip)     <-- Translation in z-direction (up/down)     }

\texttt{(  0,  0,  0, droll,  0,  0, grip)     <-- Rotation in roll axis       }

\texttt{(  0,  0,  0,  0, dpitch,  0, grip)     <-- Rotation in pitch axis  }

\texttt{(  0,  0,  0,  0,  0, dyaw, grip)     <-- Rotation in yaw axis }

\vspace{4mm}

If the grip = 100, the robot is having gripper closed. if the grip = -100, the robot is having gripper open.

Note on the gripper:

The robot's gripper should be closed if it is beginning to grasp the object, or when it is holding the object. 

When it is approaching the object, the gripper is open.

If the robot gripper needs to be closed, you should continue to close the gripper, even if it is closed.

Similarly, if the robot gripper needs to be open, you should continue to open the gripper, even if it is already open.
            
You receive the following human language correction at the current state. Pay close attention to the human language correction, interpret the human intention, and use it to arrive at the solution.

\vspace{4mm}

Some pointers for human language correction interpretation:

Move backward: decrease the x position

Move forward: increase the x position

Move left: decrease the y position

Move right: increase the y position

Move up: increase the z position

Move down: decrease the z position

Rotate: changing the yaw angle

\vspace{4mm}

Human language correction:

\begin{verbatim}
{} # to be filled in
\end{verbatim}

\vspace{4mm}

\color{black}
Consider: Does this correction concern about the gripper state (open / close)? Answer true or false. Do not care about the other action dimensions first.

If it is true, does the human want the gripper be open (grip = -100) or close (grip = 100)?

    \end{myquote}
    \caption{Prompt for selecting the gripper state.}
\end{quotebox}

\vspace{-1mm}
\begin{quotebox}[H]
    \begin{myquote}
Put the above result in format of json. You must follow the json format!

If there is no change, return:
\{``grip'': null\}

If should be open, return:
\{``grip'': -100\}

If should be close, return:
\{``grip'': 100\}
    \end{myquote}
    \caption{Prompt for putting the gripper state in json format.}
\end{quotebox}

\subsection{Example Output for Gripper State Selection}

\subsection*{\textbf{Example 1}}

Language Correction: 

\begin{verbatim}
    "You should move the gripper slightly to the left to aim it at the pea can"
\end{verbatim}

LLM Output: 
\begin{llmoutput}[H]
    \begin{llm_quote}
Based on the human language correction:

"You should move the gripper slightly to the left to aim it at the pea can"

Does this correction concern gripper state open/close? 

Answer: false.

The provided correction does not explicitly provide any instruction about changing the gripper's state (open/close); it is about the movement of the gripper in the y-axis direction (left).

If it were true, an explanation for whether the gripper should be open or close would be necessary, but since the statement is false, we need not evaluate the gripper's state from this correction. The action to be performed according to the provided correction is a slight movement to the left (which translates to: decrease the y position).

So, based on the OSC\_POSE controller manual, an appropriate action vector to realize this instruction, not considering other dimensions, would look something like this:

(0, dy, 0, 0, 0, 0, grip)

Where "dy" is a small negative value to indicate a slight move to the left and "grip" remains the same as it is not specified to change in the correction.

    \end{llm_quote}
    \caption{Results for selecting the gripper state.}
\end{llmoutput}

\subsection*{\textbf{Example 2}}

Language Correction: 

\begin{verbatim}
    "You should not release! And you should move backwards to aim at the bin."
\end{verbatim}

LLM Output: 
\begin{llmoutput}[H]
    \begin{llm_quote}
Based on the human language correction: "You should not release! And you should move backwards to aim at the bin."

Consider: Does this correction concern gripper state open/close?

Answer: True

Explanation: The phrase "You should not release!" directly pertains to the state of the gripper, indicating that the gripper should not let go of whatever it is holding.

If it is true, does the human want the gripper be open or close?

Answer: Close

Explanation: "You should not release!" implies keeping hold of an object, thus the gripper should remain closed or be closing, depending on its current state.

    \end{llm_quote}
    \caption{Results for selecting the gripper state.}
\end{llmoutput}

\subsection{Prompt and Example Output for 6D Action Selection}

\begin{quotebox}[H]
    \begin{myquote}

You are a helpful assistant who is good at employing math and computer science tools to arrive at the solution. You analyze numerical values carefully and think step by step. You will also pay close attention to the human language correction, interpret the human intention, and use it to arrive at the solution. Please describe in detail how you apply your mathematical and computational abilities, to arrive at solutions.
\vspace{4mm}
 
You have a robot arm which is the Franka Emika Panda robot arm, a single robot arm with 7 degrees of freedom.

The robot a parallel-jaw gripper equipped with two small finger pads, that comes shipped with the robot arm.

The robot comes with a controller that takes in actions. 

The expected action space of the OSC\_POSE controller (without a gripper) is \texttt{(dx, dy, dz, droll, dpitch, dyaw)}. 

The manual reads like the following: 

\texttt{( dx,  0,  0,  0,  0,  0, grip)     <-- Translation in x-direction (forward/backward) }     

\texttt{(  0, dy,  0,  0,  0,  0, grip)     <-- Translation in y-direction (left/right) }

\texttt{(  0,  0, dz,  0,  0,  0, grip)     <-- Translation in z-direction (up/down)     }

\texttt{(  0,  0,  0, droll,  0,  0, grip)     <-- Rotation in roll axis       }

\texttt{(  0,  0,  0,  0, dpitch,  0, grip)     <-- Rotation in pitch axis  }

\texttt{(  0,  0,  0,  0,  0, dyaw, grip)     <-- Rotation in yaw axis }

If the grip = 100, the robot is having gripper closed. if the grip = -100, the robot is having gripper open.

\vspace{4mm}

The robot will now perform a task. Your job is that, given a few choices of actions to perform at the current state, you will choose the correct action for the robot to perform.

\vspace{4mm}

Note on the robot position and angle:

You should consider the position and angle of the robot end effector and object, and how they are related to each other. For example, if the robot end effector is on the left of the object, you should consider moving the robot end effector to the right. If the robot end effector is not aligned with the object in rotation, you should consider rotating the robot end effector to align with the object.

\vspace{4mm}

Note on the robot gripper:

The robot's gripper should be closed if it is beginning to grasp the object, or when it is holding the object. When it is approaching the object, the gripper is open. If the robot gripper needs to be closed, you should continue to close the gripper, even if it is closed. Similarly, if the robot gripper needs to be open, you should continue to open the gripper, even if it is already open.

\vspace{4mm}
 
Given the robot and object position, first explain what stage is the task currently in, and what is the relationship between the robot and object. Explain what a good action is supposed to do.

Then based your result, look at the given actions, and return which of the following actions is the correct action to take. Let's think step by step. Explaining your reasoning before arriving at the solution. 

You always produce a single Action value in the end, which is a single number. You must follow this format!

If there are multiple actions, you must only return one of them.

\vspace{4mm}
 
You also receive a human language correction given at the current state. Pay close attention to the human language correction,
 interpret the human intention, and use it to arrive at the solution.

Some pointers for human language correction interpretation:

Move backward: decrease the x position

Move forward: increase the x position

Move left: decrease the y position

Move right: increase the y position

Move up: increase the z position

Move down: decrease the z position

Rotate: changing the yaw angle
    \end{myquote}
    \caption{Prompt for querying LLM for 6D action (shared across tasks and states).}
    \label{prompt_query}
\end{quotebox}

\begin{quotebox}[H]
    \begin{myquote}

    In this task, the robot must pick a square nut and place it on a rod. The nut has a handle to be grasped.
    
    The task has the following stages:
    
    1. Grasping the Handle: Approach the square nut's handle. 
    
    2. Peg Insertion: Lift the nut and get closer to the peg.
    
    \vspace{4mm}
    
    Input: 
        
        Robot Position: [-7  7 97]
        
        Robot Angles: [ 0 -5 -8]
        
        Handle Position: [-16  13  83]
        
        Handle Angles: [  0   0 -95]
        
        Peg Position: [23 10 85]
        
        Peg Angles: [0 0 0]

        Gripper State: [-100]

    \vspace{4mm}
    
    Action Choices:         
        
        Action 0: [  20    0    0    0    0    0 -100]
        
        Action 1: [   0   20    0    0    0    0 -100]
        
        Action 2: [   0    0   20    0    0    0 -100]
        
        Action 3: [   0    0    0    0    0   20 -100]
        
        Action 4: [ -20    0    0    0    0    0 -100]
        
        Action 5: [   0  -20    0    0    0    0 -100]
        
        Action 6: [   0    0  -20    0    0    0 -100]
        
        Action 7: [   0    0    0    0    0  -20 -100]

    \vspace{4mm}

    Human Language Correction: 

        You should not close the gripper now, you should move backwards to aim at the handle first.

    \end{myquote}
    \caption{Prompt for querying LLM for 6D action (tasks and states dependent).}
\end{quotebox}
\clearpage
Here is the corresponding LLM response for the above task, state, action and human correction information. Note that the LLM is correct about most of the reasoning except for one place -- it says the robot should move ``slightly to the left (decrease y)'' that contradicts the true spatial relationship between the robot and the handle. Nonetheless, it is able to return the correct action with the human verbal correction.

\begin{llmoutput}[H]
    \begin{llm_quote}

Let's analyze the given information step by step:

\vspace{4mm}
1. Determine the current stage of the task:

Since the gripper state is -100 (open), and the robot end effector is not close to the handle, we can conclude that the task is in the "Grasping the Handle" stage.

\vspace{4mm}
2. Analyze the relationship between the robot and the object:

The robot position is at [-7  7 97], and the handle position is at [-16  13  83]. The robot is to the right and above the handle, and it needs to move backward (decrease x) and slightly to the left (decrease y) to align with the handle.

\vspace{4mm}
3. Interpret the human language correction:

The human language correction says, "You should not close the gripper now, you should move backwards to aim at the handle first." This means we should look for an action that moves the robot backward (decrease x) and keeps the gripper open (grip = -100).

\vspace{4mm}
4. Find the correct action among the given choices:

Based on the human language correction and our analysis, we need an action that moves the robot backward (decrease x) while keeping the gripper open (grip = -100).

\vspace{4mm}
Looking at the given action choices, we can see that Action 4: [ -20    0    0    0    0    0 -100] matches our requirements. The robot moves backward (decreases x) by 20 units while keeping the gripper open (grip = -100).

\vspace{4mm}
Therefore, the correct action to take is Action 4.

\vspace{4mm}
final action idx: 4

    \end{llm_quote}
    \caption{Response for choosing the 6D action.}
\end{llmoutput}

\clearpage

\section{More Prompts}

\subsection{Prompts for Different Action Proposal Methods}

Here we provide prompts for different action proposal methods. We modify the part in Prompt \ref{prompt_query} that specifies the form of output actions. The following Prompt \ref{query_ours_onedim} is taken directly from Prompt \ref{prompt_query}.

\begin{quotebox}[H]
    \begin{myquote}
... Then based your result, look at the given actions, and return which of the following actions is the correct action to take.

Let's think step by step.
Explain your reasoning before arriving at the solution. 

You always produce a single Action value in the end, which is a single number. You must follow this format!
If there are multiple actions, you must only return one of them.
    \end{myquote}
    \caption{Prompt for ``Onedim Actions'' and ``Onedim + Original''.}
    \label{query_ours_onedim}
\end{quotebox}

\begin{quotebox}[H]
    \begin{myquote}

... Then based your result, return a correct action to take on the current state in the format of [dx, dy, dz, droll, dpitch, dyaw, grip] as mentioned above. The action value should be in the appropriate action scale (between -100 to 100).

Let's think step by step.
Explain your reasoning before arriving at the solution. 

You always produce an action being in a list of length 7. You must follow this format!

    \end{myquote}
    \caption{Prompt for ``LLM Gives Actions''.}
\end{quotebox}

\begin{quotebox}[H]
    \begin{myquote}

... Then based your result, identify the action dimension indices that requires modification. 

Then modify the original action in these action dimension indices in the appropriate action scale (between -100 to 100).

Finally, return a correct action to take on the current state in the format of [dx, dy, dz, droll, dpitch, dyaw, grip] as mentioned above.

Let's think step by step.
Explaining your reasoning before arriving at the solution. 

You always produce an action being in a list of length 7. You must follow this format!

    \end{myquote}
    \caption{Prompt for ``LLM Edits Actions''.}
\end{quotebox}

\subsection{Prompts for Summarization}

LLM generate response in a passage of reasoning process. Here we provide prompts to summarize LLM output into a standardized json form. 

\begin{quotebox}[H]
    \begin{myquote}
Now based on the previous response, summarize what is the final action choice. 

Return the answer as a JSON object, with a single key 'action', and a single value which is a number. 

Do not return any other string besides the json object. For example, if the action is 7, return {'action': 7}
If the text have multiple results for the correct action, you must only return one of them. Do not return multiple answers!
    \end{myquote}
    \caption{Summarization Prompt for ``Onedim Actions'' and ``Onedim + Original''.}
\end{quotebox}

\begin{quotebox}[H]
    \begin{myquote}
This is incorrect format. You should return the answer as single JSON object, with a single key 'action', and the value should be a single number! 

If the text have multiple results for the correct action, you must only return one of them. Do not return multiple answers! Please try again.
    \end{myquote}
    \caption{Corrective summarization Prompt for ``Onedim Actions'' and ``Onedim + Original''.}
\end{quotebox}

\begin{quotebox}[H]
    \begin{myquote}
        
Now based on the previous response, summarize what is the final action choice. 

Return the answer as a JSON object, with a single key 'action', and a single list. The value of JSON object must be a list of 7 numbers.

Do not return any other string besides the json object. 

\vspace{4mm}

For example, if the action is [0,0,20,0,0,-30,100], return {'action': [0, 0, 20, 0, 0, -30, 100]}.

If the action is [0,0,20,0,0,-30,100], return {'action': [0, 0, 20, 0, 0, -30, 100]}.

If the action is [0,20,20,0,0,0,-100], return {'action': [0, 20, 20, 0, 0, 0, -100]}.

If the action is [-20 20 0 0 0 20 100], return {'action': [-20, 20, 0, 0, 0, 20, 100]}.

If the action is [0,-20,0,0,0,0,100], return {'action': [0, -20, 0, 0, 0, 0, 100]}.

If the action is [20,0,0,0,0,0,-100], return {'action': [20, 0, 0, 0, 0, 0, -100]}.

If the action is 0 0 0 0 0 -20 -100, return {'action': [0, 0, 0, 0, 0, -20, -100]}.

If the action is [14 20 0 0 5 0 -100], return {'action': [14, 20, 0, 0, 5, 0, -100]}.

If the action is -1 0 2 -40 30 1 100, return {'action': [-1, 0, 2, -40, 30, 1, 100]}
    \end{myquote}
    \caption{Summarization Prompt for ``LLM Gives Actions'' and ``LLM Edits Actions''.}
\end{quotebox}

\begin{quotebox}[H]
    \begin{myquote}
This is incorrect format. You should return the answer as single JSON object, with a single key 'action', and the value should be a single list! Please try again.
    \end{myquote}
    \caption{Corrective summarization Prompt for ``LLM Gives Actions'' and ``LLM Edits Actions''.}
\end{quotebox}
\clearpage
\section{Policy Implementations}
\label{appendix:policy_imp}

We describe the policy architecture details initally introduced in Section \ref{sec:setup}. Our codebase is based on robomimic \cite{robomimic}, an open-source project that benchmarks a range of learning algorithms on offline data. We standardize all methods with the same state-of-the-art policy architectures and hyperparameters from robomimic. The architectural design includes a transformer policy backbone, ResNet-18 image encoders, random cropping for image augmentation, GMM head, and the same training procedures. The list of hyperparameter choices is presented in Table \ref{table:common}. We also include the task level hyperparameters (e.g., image size, number of rollouts) for simulation tasks in Table \ref{table:sim-tasks} and that for real-robot tasks in Table \ref{table:real-tasks}.

\begin{table*}[h]
\vspace{15pt}
\centering
\caption{Model training hyperparameters}
\begin{tabular}{c c}
\hline
\textbf{Hyperparameter} & \textbf{Value}\\
\hline
& \\

Context length & $10$ \\
Embedding dim & $512$ \\
Num of layers & $6$ \\
Num of heads & $8$ \\
Embedding dropout & $0.1$ \\
Attention dropout & $0.1$ \\
Block output dropout & $0.1$ \\
Activation & gelu \\

& \\

GMM number of modes & $5$\\
Image encoder & ResNet-18\\
Random crop ratio & $90$\% of image height \\
& \\
Optimizer & Adam\\
Batch size & $16$\\

\# Training steps per epoch & $500$\\
\# Total training epochs & $1000$ \\
& \\
Evaluation checkpoint interval (in epoch) & $100$ \\
& \\

\hline
\end{tabular}
\label{table:common}
\end{table*}

\begin{table*}[h]
\vspace{15pt}
\centering
\caption{Task hyperparameters (simulation)}
\begin{tabular}{c c c c c}
\hline
\textbf{Hyperparameter} & \texttt{Pick Place Can} & \texttt{Threading} & \texttt{Square} & \texttt{Coffee Machine} \\
\hline
& \\
Workspace camera image size ($h \times w$) & $84 \times 84 $ & $84 \times 84$ & $84 \times 84$ & $84 \times 84$ \\
Wrist camera image size ($h \times w$) & $84 \times 84 $ & $84 \times 84$ & $84 \times 84$ & $84 \times 84$ \\
\# human demonstrations & $50$ & $50$ & $50$ & $50$ \\
\# rollouts in interaction & $50$ & $50$ & $50$ & $50$ \\
Evaluation rollout length & $300$ & $500$ & $400$ & $450$ \\
& \\

\hline
\end{tabular}
\label{table:sim-tasks}
\end{table*}

\begin{table*}[h]
\vspace{15pt}
\centering
\caption{Task hyperparameters (real robot)}
\begin{tabular}{c c c }
\hline
\textbf{Hyperparameter}  & \texttt{PickPlace-Bin} & \texttt{PickPlace-Drawer-Basket} \\
\hline
& \\
Workspace camera image size ($h \times w$) & $84 \times 84$ & $84 \times 112$ \\
Wrist camera image size ($h \times w$) & $84 \times 84$ & $84 \times 84$ \\
\# human demonstrations & $40$ & $40$ \\
\# rollouts in interaction & $80$ & $80$ \\
Evaluation rollout length & $1000$ & $1000$ \\
& \\

\hline
\end{tabular}
\label{table:real-tasks}
\end{table*}

\end{document}